\pdfoutput=1

\documentclass[11pt]{article}
\usepackage{acl}

\usepackage{times}
\usepackage{latexsym}

\usepackage[T1]{fontenc}
\usepackage[utf8x]{inputenc}

\usepackage{microtype}

\usepackage{inconsolata}

\usepackage{graphicx}
\usepackage{amsmath}
\usepackage{booktabs}
\usepackage{multirow}
\usepackage{caption}
\usepackage{subcaption}

\usepackage{amssymb, upgreek}
\usepackage{color}
\definecolor{keywordcolor}{rgb}{0.7, 0.1, 0.1}   
\definecolor{commentcolor}{rgb}{0.4, 0.4, 0.4}   
\definecolor{symbolcolor}{rgb}{0.0, 0.1, 0.6}    
\definecolor{sortcolor}{rgb}{0.1, 0.5, 0.1}      
\definecolor{errorcolor}{rgb}{1, 0, 0}           
\definecolor{stringcolor}{rgb}{0.5, 0.3, 0.2}    
\usepackage{listings}

\newcommand{\method}{\textit{Safe}}
\newcommand{\bench}{\textit{FormalStep}}

\DeclareMathOperator*{\argmax}{arg\,max}

\title{\method: Enhancing Mathematical Reasoning in Large Language Models\\ via Retrospective \textit{S}tep-\textit{a}ware \textit{F}ormal V\textit{e}rification}

\author{
 \textbf{Chengwu Liu\textsuperscript{1}$^*$},
 \textbf{Ye Yuan\textsuperscript{1}},
 \textbf{Yichun Yin\textsuperscript{2}},
 \textbf{Yan Xu\textsuperscript{2}},
 \textbf{Xin Xu\textsuperscript{3}$^*$},
\\
 \textbf{Zaoyu Chen\textsuperscript{4}$^*$},
 \textbf{Yasheng Wang\textsuperscript{2}},
 \textbf{Lifeng Shang\textsuperscript{2}},
 \textbf{Qun Liu\textsuperscript{2}},
 \textbf{Ming Zhang\textsuperscript{1}},
\\
\\
 \textsuperscript{1}School of Computer Science, National Key Laboratory for Multimedia Information Processing, \\PKU-Anker LLM Lab, Peking University,
 \textsuperscript{2}Huawei Noah’s Ark Lab,
\\
 \textsuperscript{3}The Hong Kong University of Science and Technology,
 \textsuperscript{4}The Hong Kong Polytechnic University,
\\
 \small{
   \textbf{Correspondence:} \href{mailto:mzhang_cs@pku.edu.cn}{Ming Zhang (\texttt{mzhang\_cs@pku.edu.cn})}
 }
}

\newif\ifshowcomment
\showcommenttrue

\ifshowcomment
\newcommand{\yuanye}[1]{\textcolor{orange}{[Yuanye: #1]}}
\newcommand{\todo}[1]{\textcolor{red}{[TODO: #1]}}
\else
\newcommand{\yuanye}[1]{}
\newcommand{\todo}[1]{}
\fi

\begin{document}
\maketitle
\def\thefootnote{*}\footnotetext{Work done during the internship at Huawei Noah’s Ark Lab. The code and dataset are available at \url{https://github.com/liuchengwucn/Safe}.}
\def\thefootnote{\arabic{footnote}}

\begin{abstract}
Chain-of-Thought (CoT) prompting has become the de facto method to elicit reasoning capabilities from large language models (LLMs).
However, to mitigate hallucinations in CoT that are notoriously difficult to detect, current methods such as process reward models (PRMs) or self-consistency operate as opaque boxes and do not provide checkable evidence for their judgments, possibly limiting their effectiveness.
To address this issue, we draw inspiration from the idea that ``the gold standard for supporting a mathematical claim is to provide a proof'' \cite{avigad2021theorem}.
We propose a retrospective, step-aware formal verification framework \method{}. Rather than assigning arbitrary scores, we strive to articulate mathematical claims in formal mathematical language Lean 4 at each reasoning step and provide formal proofs to identify hallucinations.
We evaluate our framework \method{} across multiple language models and various mathematical datasets, demonstrating a significant performance improvement while offering interpretable and verifiable evidence.
We also propose \bench{} as a benchmark for step correctness theorem proving with $30,809$ formal statements.
To the best of our knowledge, our work represents the first endeavor to utilize formal mathematical language Lean 4 for verifying natural language content generated by LLMs, aligning with the reason why formal mathematical languages were created in the first place: to provide a robust foundation for hallucination-prone human-written proofs.

\end{abstract}

\section{Introduction}

The practice of guiding LLMs to generate additional chain of thought during inference has emerged as a paradigm for enhancing the reasoning capabilities of these models \citep{wei2022chain, NEURIPS2022_8bb0d291, metamath2023yu,xu2024egsm}.
While it is feasible to generate them on a large scale with manageable costs, verifying the absence of hallucinations in these steps remains challenging.
The verification process is especially crucial in domains such as mathematical reasoning~\citep{llm4math2024Ahn, xu2025ugmathbench}, code generation~\citep{zhuo2024bigcodebench,li2024evocodebench}, and many others~\citep{SciBench2023Wang, gpqa2023rein}, where even minor errors can significantly disrupt subsequent generations and the final outcome\citep{shen2021generate, zelikman2022star}.
Consequently, verifying the accuracy of each individual step becomes crucial to the overall performance.

To mitigate hallucinations, a prevalent strategy involves employing a verification mechanism that evaluates and assigns scores to the reasoning trajectories generated by LLMs \citep{lightman2024lets, wang2024mathshepherd, xu2024pds}.
This approach facilitates the selection of the most promising responses from a pool of candidates, commonly referred to as Best-of-N (BoN) sampling \citep{shen2021generate, cobbe2021training}.
Although effective, these approaches perceive the LLM-based verifier as an opaque box, thereby forfeiting the benefits of symbolic computation and formal verification.
Consequently, they inherently lack interpretability and do not provide guarantees of correctness.

\begin{figure}[t]
  \centering
  \includegraphics[width=\columnwidth]{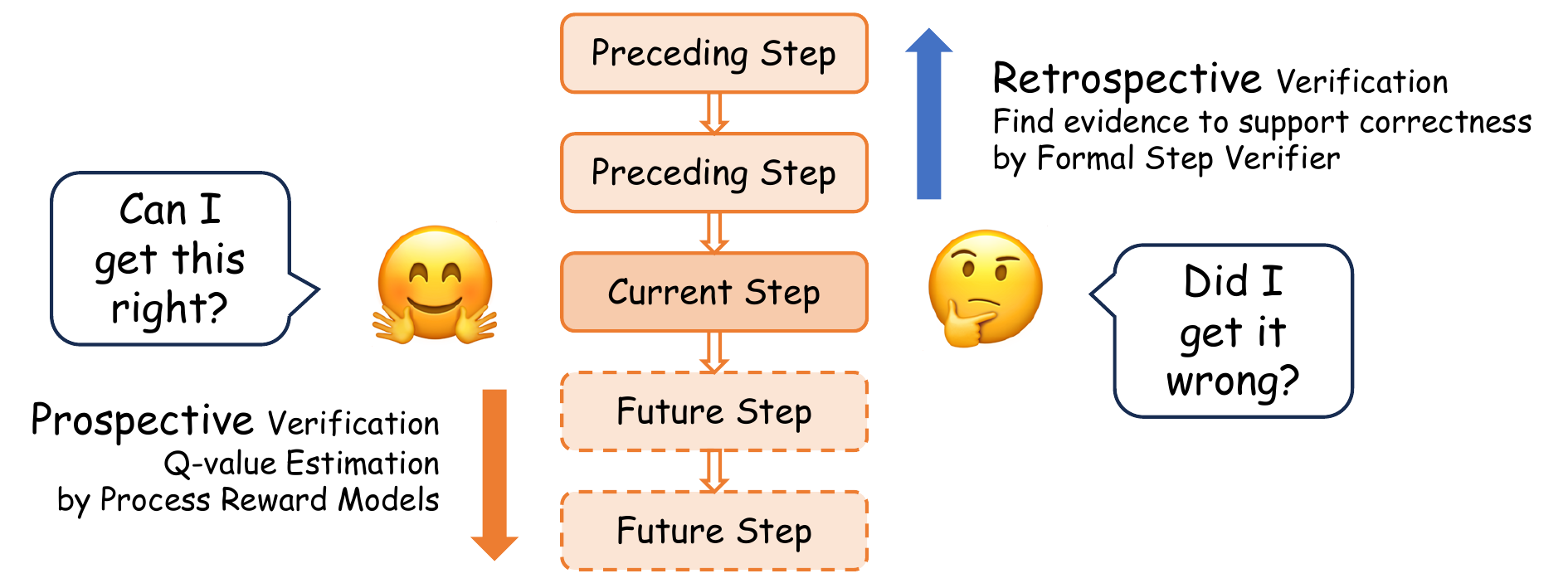}
  \caption{The distinction between prospective verification and retrospective verification.
  }
  \label{fig:per_retro}
\end{figure}

In this study, we propose a formal verification framework designed to enhance the strengths of natural language reasoning, which is characterized by its abundant data and diverse reasoning forms \citep{huang2022towards}, with formal language reasoning, known for its ability to verify correctness and provide better interpretability \citep{pan2023logic}.
During the inference phase of LLMs, we decompose complex mathematical reasoning trajectories, which are challenging to verify entirely, into a series of simpler steps.
For each step, we utilize an LLM to automatically generate formal statements that substantiate the correctness of that particular step, akin to the process of auto-formalization.
These formal statements necessitate only the evaluation of single-step correctness, allowing them to be effectively addressed by readily available automated theorem provers.
By aggregating the formal statements and proofs from each step as evidence, we can evaluate the overall reasoning trajectory by scoring its state sequences.

Note that our formal verifier focuses on the correctness of each step, which constitutes retrospective verification.
In contrast, PRMs are typically trained using a loss function that evaluates the likelihood of achieving a correct outcome in the future \citep{hao2023reasoning, wang2024mathshepherd}, a process that can be perceived as prospective verification.
The distinction between these two approaches is illustrated in Figure~\ref{fig:per_retro}.
Our proposed approach \method{}, integrates retrospective formal scores with those of a prospective PRM, resulting in significant performance improvement.
This outcome highlights the potential of combining formal reasoning and natural language reasoning, an approach commonly referred to as the neuro-symbolic system \citep{besold2021neural, sarker2022neuro}.

In conclusion, we present the following three main contributions:
(1) We propose a novel auto-formalization task, which aims to generate formal statements that validate the correctness of one specific step instead of simply translating the problem. Our dataset comprising $30,809$ formal statements, referred to as \bench{}, will be released to facilitate auto-formalization and automated theorem proving in low compute settings.
Our empirical findings indicate that, despite these statements being out of distribution, they can still be effectively addressed by off-the-shelf automated theorem provers, provided there is an adequate computational budget.
(2) We propose a step-level formal verifier that outputs one of four distinct states, depending on whether a given step can be auto-formalized by an LLM and resolved by an automated theorem prover, rather than providing numerical scores for each step.
To the best of our knowledge, this is the first study utilizing formal mathematics language Lean 4 to verify the correctness of mathematical reasoning trajectories generated by LLMs.
(3) We propose a formal verification framework \method{} that aggregates a sequence of states generated by the formal step verifier.
Furthermore, we demonstrate that the retrospective scores generated by our formal verifier can be effectively integrated with the prospective scores, resulting in state-of-the-art performance.

\begin{figure*}[t]
  \centering
  \includegraphics[width=2\columnwidth]{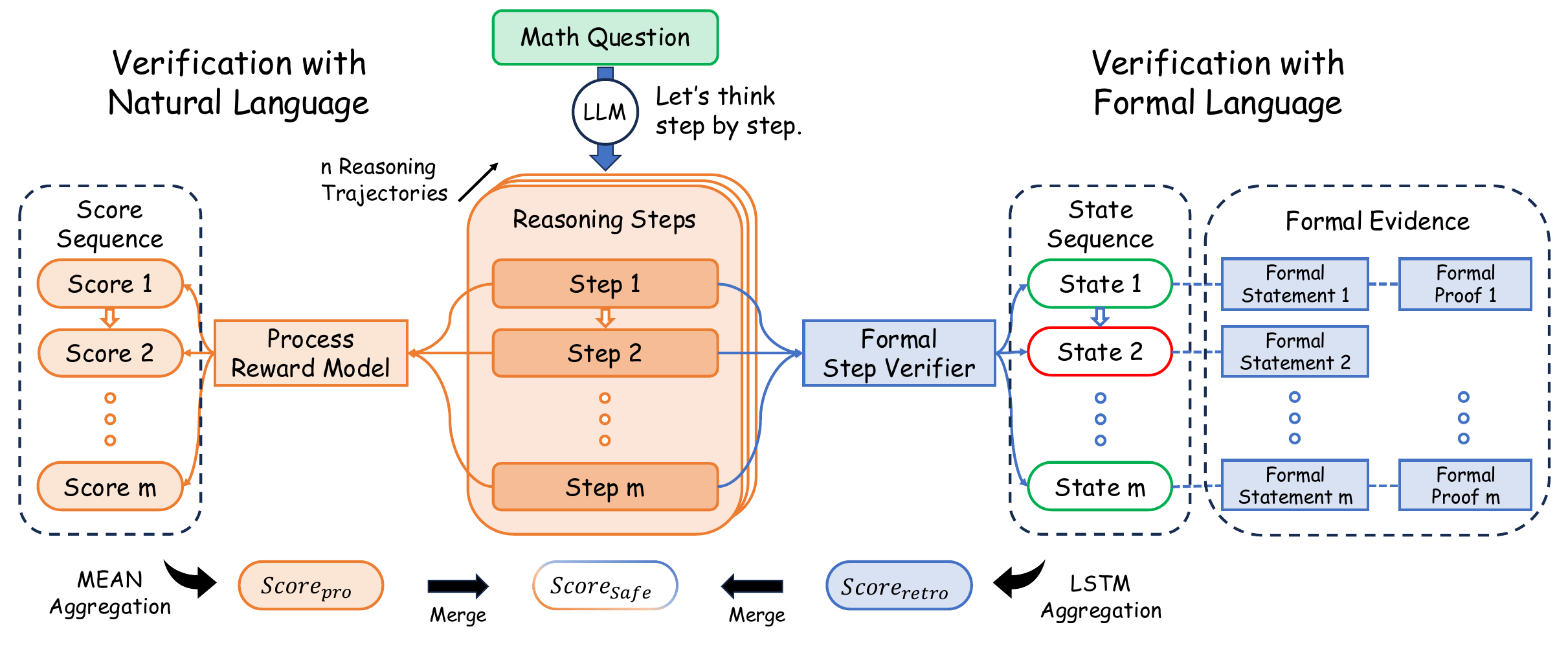}
  \caption{The primary pipeline flowchart. Solutions for each mathematical problem, generated using the zero-shot CoT Prompt, are decomposed into reasoning steps. Each step is evaluated by both a Formal Verifier and a process reward model (PRM), which assesses the state or score of each step. Following this evaluation, the states or scores are aggregated to yield retrospective and prospective scores. These two scores can subsequently be combined to generate a final evaluation score.}
  \label{fig:main_pipeline}
  \vspace{-0.5cm}
\end{figure*}

\section{Related Work}

The rapid advancement of LLMs has catalyzed transformative applications across diverse domains, ranging from medical diagnostics to fundamental scientific discovery~\citep{jumper2021highly,jumper2020alphafold2,madani2023large,yang2024poisoning,feng2024bioactivity,shao2024deepseekmath,ying2024internlm}. Particularly in the realm of mathematical reasoning, which serves as a critical benchmark for evaluating artificial intelligence systems, researchers have embarked on two distinct yet complementary research trajectories. The first strand focuses on formal mathematics, where scholars investigate how LLMs can assist in constructing machine-verifiable mathematical proofs through interactive theorem provers like Lean and Coq~\citep{polu2020generative,wang-etal-2023-dt,xin2024deepseek1,xin2024deepseek15}. Concurrently, a parallel research effort examines the models' capacity for solving mathematical word problems expressed in natural language, aiming to develop general-purpose systems capable of parsing complex problem statements, generating stepwise solutions, and providing rigorous mathematical justifications~\citep{NEURIPS2022_8bb0d291,shao2024deepseekmath,ying2024internlm}. 
This bifurcation in research directions reflects both the multifaceted nature of mathematical intelligence and the evolving capabilities of modern language models.

\subsection{Automated Theorem Proving}
The objective of automated theorem proving (ATP) is to produce a formal proof process composed of a sequence of tactics for statements articulated in a formal language.
This proof process is designed to be automatically verifiable by a machine, thereby ensuring its correctness.
Pioneering work in this field GPT-f is trained using the Metamath \texttt{set.mm} dataset and employs a best-first search method to iteratively generate formal theorem proofs \citep{polu2020generative}.
Subsequent research primarily focuses on implementing improved search strategies \citep{lample2022hypertree, wang-etal-2023-dt}, leveraging the capabilities of formal theorem proving environments into the proving process \citep{yang2023leandojo, thakur2024context, poluformal}, and synthesizing new data to enhance the quality of the training dataset \citep{hanproof, xin2024deepseek1, huangmustard}.
The benchmarks for automated theorem proving include \texttt{miniF2F}\citep{minif2f}, \texttt{FIMO} \citep{liu2023fimo}, \texttt{TRIGO}~\citep{xiong2023trigo}, among others.

Despite significant advancements in ATP through the synthesis of new data via expert iteration and the enhancement of model exploration within the solution space using complex tree search techniques, solving complex mathematical problems remains a challenging and computationally intensive task.
For example, despite a sample budget of $16 \times 6400$, DeepSeek-Prover achieves a $60.2\%$ accuracy rate on the \texttt{miniF2F} benchmark~\citep{minif2f}, which includes formalized problems from high school, competition, and undergraduate mathematics.
Additionally, it attains merely a $3.4\%$ accuracy on the \texttt{FIMO} dataset, which consists of more challenging problems typical of the International Mathematical Olympiad~\citep{xin2024deepseek15}.

\subsection{Auto-Formalization}
ATP represents a particularly challenging task, primarily due to the limited availability of formal mathematical data \cite{wu2022autoformalization}.
Moreover, the costs associated with employing domain experts for annotation are prohibitively high.
To address this issue, auto-formalization leverages the in-context learning capabilities of LLMs to transform abundant preexisting mathematical data into formal data \citep{wu2022autoformalization, lu2024process, jiang2024leanreasoner}.

Currently, existing efforts in auto-formalization are focusing on natural language mathematical datasets created by humans \citep{jiangdraft, ying2024lean}, as these datasets are already abundant and of high-quality \citep{DBLP:journals/corr/abs-1910-09336}. These efforts primarily focus on the formalization of mathematical statements rather than their proofs.

\subsection{Process Reward Models}
In the fields of mathematical reasoning and code generation, existing research suggests that process reward models (PRMs) are more effective than outcome reward models (ORMs) \citep{lightman2024lets, wang2024mathshepherd}.
An ORM assesses the overall performance of the whole output of LLMs, while a PRM evaluates each individual step, providing more fine-grained feedback.

PRMs have two primary applications \citep{wang2024mathshepherd}.
First, they can be employed during the post-training reinforcement learning phase, where an LLM samples its outputs and learns from trajectories that receive higher scores from a PRM, thereby facilitating self-improvement \citep{shao2024deepseekmath, lai2024step, yan2024s3cmath}.
Second, PRMs can be used during the inference phase of LLMs, allowing the model to sample multiple responses and select the Best-of-N as the final output.
Numerous studies have shown that this approach can enhance the performance of LLMs in reasoning tasks \citep{havrillaglore, setlurrewarding}.

\subsection{Neuro-symbolic AI}

In the domains of formal language mathematical reasoning and natural language mathematical reasoning, while these two approaches differ methodologically, they possess complementary and mutually reinforcing capabilities that can collectively advance the frontier of mathematical reasoning in artificial intelligence.
This line of research is often referred to as Neuro-symbolic AI, which integrates the strengths of neural networks and symbolic logic-based reasoning to develop more robust and intelligent systems \citep{yang2024formal}.

Draft, Sketch and Prove represents a representative approach that employs natural language to augment formal language reasoning \citep{jiangdraft}. This methodology primarily focuses on automated theorem proving, where natural language serves as supplementary guidance to enhance the success rate of formal theorem provers.
On the other hand, LINC \citep{olausson2023linc} and Logic-LM \citep{pan2023logic} represent prominent approaches that leverage formal languages to enhance natural language reasoning.
These methods primarily employ automatic formalization to transform natural language tasks into formal representations before solving them through symbolic reasoning.
Besides, DTV employs Isabelle to formally verify the quantitative reasoning capabilities of LLMs \citep{zhou2024don}.
\section{Methodology}

For any given mathematical problem $P$, we employ zero-shot chain-of-thought (CoT) prompting to sample $n$ output results from an LLM, denoted as $A_1, A_2, \ldots, A_n$.
Each output $A_i$ is decomposed into a sequence of steps represented by:
\begin{equation*}
    A_i = \texttt{concat}(step_{i1}, step_{i2}, \ldots, step_{im_i})
\end{equation*}
, where \texttt{concat} denotes the concatenation function, and $m_i$ indicates the number of steps contained within $A_i$.
Each step undergoes formal validation through a step verifier, resulting in a verification state defined as:
\begin{equation*}
    state_{ij} = \texttt{step\_verifier}(step_{ij})
\end{equation*}
for $ i = 1, 2, \ldots, n$, $j = 1, 2, \ldots, m_i$.
The step verifier integrates two modules: auto-formalization and automated theorem proving.
Verification of these steps can be conducted concurrently to minimize end-to-end latency.
We aggregate the verification states of each step using an aggregator to produce a retrospective $score_{retro}$.
Subsequently, we utilize an off-the-shelf PRM to obtain a prospective $score_{pro}$, which is then integrated into $score_{retro}$ to derive the final score.
A diagram for the overall reasoning model is presented in Figure~\ref{fig:main_pipeline}.
The following sections will elaborate on the processes of sampling and decomposition, the auto-formalization and theorem proving conducted by the step verifier, and the details regarding the score aggregator.

\subsection{Sampling and Decomposition}

We use the zero-shot CoT prompting technique to sample $n$ outputs, denoted as:
\begin{equation*}
A_i \sim \texttt{LLM}(P, cot\_prompt) \quad \text{for} \quad i = 1, 2, \ldots, n
\end{equation*}
where the prompt is expressed in its simplest form as ``Let’s think step by step'' \citep{NEURIPS2022_8bb0d291}.

To decompose the output $A_i$ into discrete steps, we have explored two approaches: heuristic rules and LLM in-context learning (ICL) \cite{brown2020language}. 
The details of the decomposition can be found in Appendix~\ref{sec:decomposition}.

\subsection{Step Verifier}

The step verifier validates reasoning steps by formalizing natural language expressions into formal statements that establish the mathematical soundness of these steps.
We employ readily available automated theorem provers to attempt to prove these statements.
Following the proof attempt, a verification state is generated.

There are certain scenarios in which the auto-formalization process is unnecessary or impractical, which typically involve straightforward steps such as chanting, repeating, and summarizing.
Additionally, some steps are sufficiently complex and fall outside the typical scope of our formal mathematical language, Lean 4 \citep{moura2021lean}.
While Lean 4 can effectively express relationships relevant to number theory and algebra, it may face limitations in articulating concepts related to geometry and combinatorial mathematics.

Our step verifier can produce one of four possible states: 1) no verification required; 2) failed formalization, which may result from either the limitations of the Lean 4 language or the constraints of our auto-formalization pipeline; 3) successful formalization accompanied by a proof of the statement; and 4) successful formalization, but with a failure in theorem proving. The following subsections will offer a detailed description of the two essential components of our step verifier: auto-formalization and automated theorem proving.

\begin{figure*}[t]
  \includegraphics[width=0.95\linewidth]{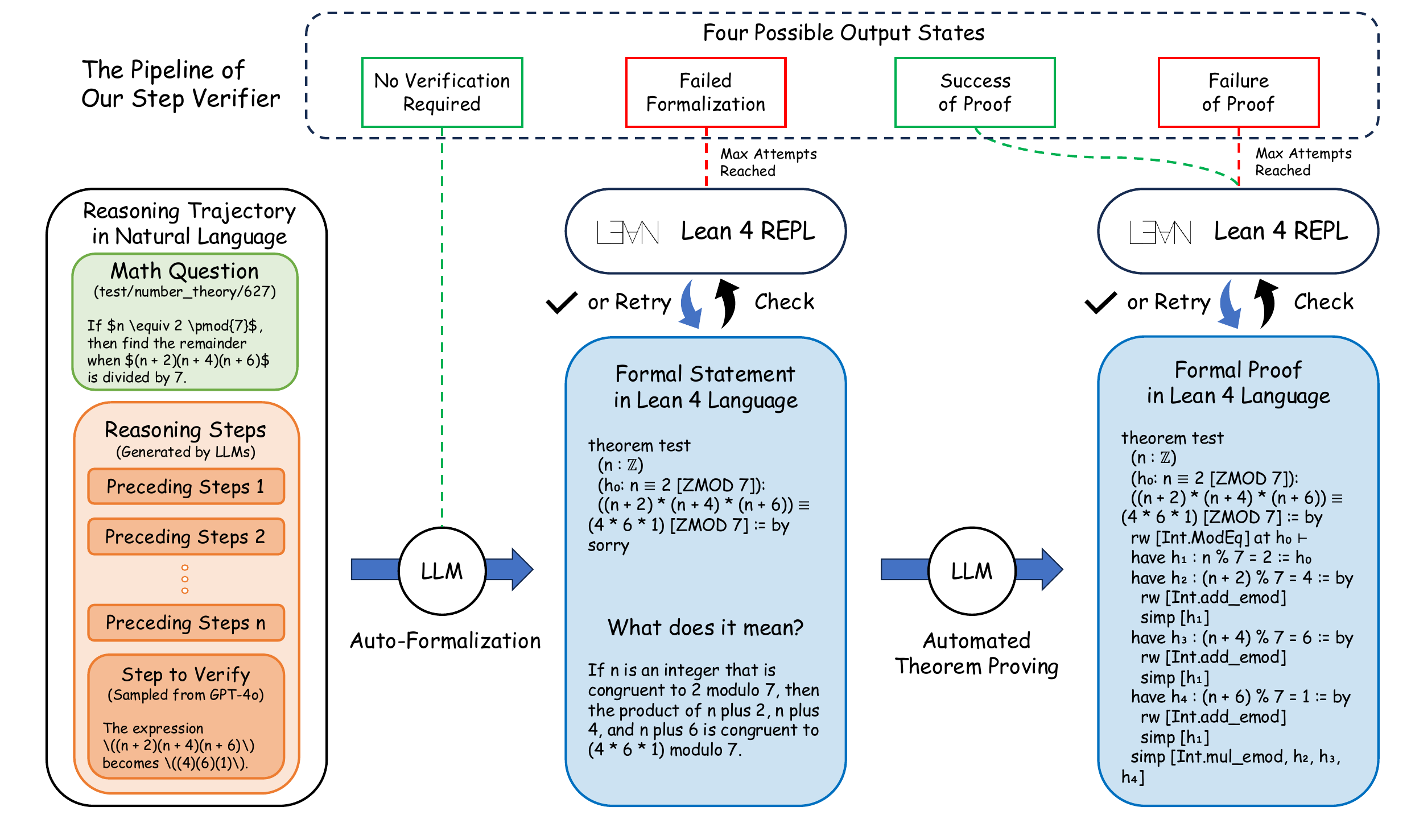}
  \centering
  \caption{The pipeline of the step verifier. We construct a prompt that incorporates the relevant question and context, then utilize the in-context learning capability of large language models to perform auto-formalization. The formalized Lean 4 theorem is then validated using the Lean REPL. Once the auto-formalization is successful, we attempt to prove the theorem using an LLM-based prover, followed by verification through the Lean REPL. Upon completion of the entire procedure, each step corresponds to one of four distinct states.}
  \label{fig:step_verifier}
  \vspace{-0.4cm}
\end{figure*}

\subsubsection{Auto-Formalization Module}
Previous efforts in auto-formalization have primarily focused on translating mathematical statements articulated in natural language into their equivalent formal mathematical statements \citep{wu2022autoformalization}.
In this study, we utilize auto-formalization in a novel manner to verify the mathematical soundness of each step.
For instance, when the reasoning LLM attempts to transform inequalities, we use the inequality prior to the transformation as a premise and the resulting inequality as the proof objective for the Lean 4 statement.
An illustrative example of auto-formalization is presented in the left half of Figure~\ref{fig:step_verifier}.
Additional examples of our auto-formalization can be found in Appendix~\ref{sec:example}.

In accordance with the existing literature \citep{wu2022autoformalization, liu2023fimo}, we leverage the capabilities of LLMs in in-context learning to automate the formalization process for our verifier.
The generated formal statements are subsequently submitted to the Lean read-eval-print-loop (REPL) environment for validation, ensuring adherence to the Lean 4 syntax.
We manually curate a selection of few-shot examples to guide the LLM in the process of auto-formalization.
Furthermore, we include instructions within the prompts to prompt the model to recognize that certain steps do not require validation or may exceed the capabilities of Lean 4. The complete prompt utilized in the auto-formalization process is available in Appendix~\ref{sec:prompt}.

\subsubsection{Automated Theorem Proving Module}

In this study, we utilize existing automated theorem provers to generate proofs for formalized statements, thereby providing evidence for the correctness of the natural language steps involved.
Specifically, we examine the performance of two state-of-the-art LLM-based automated theorem provers, COPRA \citep{thakur23language} and DeepSeek-Prover-V1.5 \citep{xin2024deepseek15}.

As previously noted, the difficulty of formally proving the validity statement of a single step is intuitively lower than that of solving an entire problem.
Consequently, while the formal statements derived from individual natural language steps may be out of distribution, they may nonetheless fall within the capabilities of contemporary automated theorem provers.
Subsequent experiments demonstrate that for the ATP task, sampling a limited number of proofs can yield a success rate exceeding 80\%.
The right half of Figure~\ref{fig:step_verifier} provides an example.

\subsection{State Aggregator}

Auto-formalization pipelines and automated theorem provers are fallible, leading to proof failures that can arise from two extra sources: (1) the auto-formalization pipeline may produce a plausible but erroneous and consequently unprovable statement despite the step being correct; or (2) while the statement may be accurate and provable, the automated theorem prover may not be able to complete the proof within a constrained sample budget.
As a result, the four states provided by the step verifier are inherently susceptible to noise.
To mitigate the effects of noise, we aggregate all the states at each step to compute one final score.

To perform the score prediction task, we utilize a tiny LSTM model \citep{hochreiter1997long}, with a token vocabulary size of $4$.
Our choice of LSTM is motivated by its simplicity and effectiveness for this specific task.
Since the step verifier outputs a sequence of four discrete states (treated as tokens), an LSTM provides an intuitive yet efficient way to model this sequential evidence.
The latent variable from the final step undergoes a linear transformation, followed by the application of the sigmoid function to yield a score within the range of 0 to 1.

\begin{align*}
   score_{retro}^i &= \sigma(W \cdot \text{LSTM}(state_{i1}, state_{i2}, \\ & \quad \ldots, state_{ij}) + b) 
\end{align*}
Note that the information derived from formal verification is highly condensed and retrospective, as each state conveys only two bits of information.
This contrasts with the prospective scores of LLM-based PRM, which uses complete reasoning steps in natural language as input to predict a Q-value, indicating whether these steps can potentially lead to the correct answer \citep{wang2024mathshepherd}.
\begin{equation*}
score_{pro}^i = \text{PRM}(step_{i1}, step_{i2}, \ldots, step_{ij}) 
\end{equation*}
Recognizing the complementarity between the two scores, we further combine the retrospective score from the state aggregator with the prospective score from an existing PRM to generate an ensemble score. We experimented with several methods for calculating ensemble scores, the detailed discussion of which can be found in Appendix~\ref{sec:ensembling}.

\begin{align*}
score_i &= {score_{retro}^i}^{\alpha} \cdot {score_{pro}^i}^{(1-\alpha)} \\
A^* &= A_{i^*}\text{, where } i^* \in \argmax_{i} score_i
\end{align*}
where $\alpha$ is a hyper-parameter.

\begin{table*}[ht]
    \centering
\begin{footnotesize}
\begin{tabular}{*{13}{c}}
  \toprule
  & \multicolumn{3}{c}{\textbf{Llama 3.1}} & \multicolumn{3}{c}{\textbf{Llama 3.0}} & \multicolumn{3}{c}{\textbf{GPT-4o}} & \multicolumn{3}{c}{\textbf{Deepseek-Math}} \\
  \cmidrule(lr){2-4}\cmidrule(lr){5-7}\cmidrule(lr){8-10}\cmidrule(lr){11-13}
  & MATH & G8K & CM & MATH & G8K & CM & MATH & G8K & CM & MATH & G8K & CM \\
  \midrule
  ZS-CoT@1 & 49.1 & 85.4 & 52.6 & 26.1 & 79.9 & 31.3 & 76.9 & 95.0 & 73.4 & 40.8 & 80.1 & 48.4 \\
  Majority@5 & 50.5 & 87.8 & 54.3 & 24.8 & 80.7 & 29.6 & 78.9 & 95.7 & 73.9 & 39.3 & 81.7 & 48.7\\
  Skywork (ORM) & 48.9 & \underline{90.2} & 53.2 & 30.5 & 76.0 & 36.6 & 76.7 & 88.6 & 73.1 & 43.6 & 76.8 & 50.7 \\
  ArmoRM (ORM) & 55.1 & 90.0 & 57.1 & 32.3 & 86.1 & 37.3 & 79.3 & 95.5 & \textbf{74.7} & 48.6 & 86.6 & 53.5 \\
  Shepherd (PRM) & \underline{58.1} & \underline{90.2} & \underline{58.3} & \underline{34.7} & \underline{86.4} & \underline{40.4} & \underline{79.8} & \underline{95.8} & 73.5 & \underline{49.7} & \underline{87.1} & \underline{55.0} \\
  RLHFlow (PRM) & 51.7 & 89.9 & 53.6 & 29.6 & 86.2 & 36.5 & 78.7 & 95.3 & \underline{74.2} & 44.4 & 86.3 & 50.9 \\
  \midrule
  \textbf{LSTM (Ours)} & 55.1 & 88.7 & 55.9 & 33.0 & 84.3 & 36.6 & 78.9 & 95.5 & 73.8 & 48.2 & 83.3 & 51.1 \\
  \textbf{\method{} (Ours)} & \textbf{60.0} & \textbf{90.8} & \textbf{59.0} & \textbf{36.3} & \textbf{87.4} & \textbf{43.4} & \textbf{80.4} & \textbf{96.0} & \underline{74.2} & \textbf{52.4} & \textbf{87.6} & \textbf{55.4} \\
  \midrule
  Pass@5 & 70.8 & 95.5 & 72.3 & 48.9 & 92.5 & 52.9 & 87.8 & 97.0 & 81.5 & 62.6 & 92.1 & 67.1 \\
  \bottomrule
\end{tabular}
\end{footnotesize}
    \caption{The experimental results of various models on the \texttt{MATH-500}, \texttt{GSM8K} and \texttt{CollegeMath} datasets are presented, denoted as MATH, G8K and CM, respectively.
    LSTM and \method{} denote our proposed methodologies.
    Pass@5 indicates the probability of correct for at least one of the five samples, representing the performance upperbound across all Best-of-N strategies.}
    \label{tab:main}
\end{table*}

\section{Experiment}

\subsection{Experimental Setup}

\noindent\textbf{Datasets and LLMs}
We evaluate BoN@5 accuracy on the \texttt{GSM8K} \cite{cobbe2021training}, \texttt{MATH-500} \citep{hendrycksmath2021, lightman2024lets}, and \texttt{CollegeMath} \citep{tangmathscale} datasets, which encompass grade school, high school, and college-level mathematics.
Our experiments includes four language models: Llama-3-8B-Instruct \citep{dubey2024llama}, Llama-3.1-8B-Instruct \citep{dubey2024llama}, gpt-4o-2024-08-06 \citep{hurst2024gpt}, and deepseek-math-7b-instruct \citep{shao2024deepseekmath}.
This selection represents a range of models from different model families and varying capacities.
We utilized GPT-4o as the LLM to perform the auto-formalization process.

\noindent\textbf{Baselines and Metrics}
We evaluate the performance in comparison to the zero-shot CoT method and the self-consistency majority voting strategy, as well as other reward models.
We choose reward models as baselines which possess a parameter size comparable to that of our automated theorem prover, the DeepSeek-Prover-V1.5 with 7B parameters.
The baseline models include both ORMs and PRMs.
We incorporate the current SOTA reasoning ORMs in RewardBench \citep{lambert2024rewardbench}, featuring Skywork-Reward-Llama-3.1-8B-v0.2 \citep{liu2024skywork} and ArmoRM-Llama3-8B-v0.1 \citep{ArmoRM} as representative examples.
Regarding PRMs, existing works are primarily domain-specific. Thus we include two recent PRMs that concentrate on mathematical reasoning, namely math-shepherd-mistral-7b-prm \citep{wang2024mathshepherd} and RLHFlow/Llama3.1-8B-PRM-Deepseek-Data \citep{xiong2024rlhflowmath}, as the representative models.

\noindent\textbf{State Aggregator Data Collection}
We randomly select a subset of the training dataset that has approximately the same size as the test dataset to train the LSTM model.
Specifically, the selection comprise $500$ problems from the \texttt{MATH} dataset, $1,000$ problems from the \texttt{GSM8K} dataset, and the entire training set from the \texttt{CollegeMath} dataset.
We query the reasoning LLM to answer these problems and the outputs are compared with the ground truth to assess the accuracy of the generated reasoning trajectories.
The sampled reasoning trajectories are subsequently utilized to train the LSTM model.

\noindent\textbf{State Aggregator Training Setup}
The LSTM model utilized in this study features a tiny number of parameters, comprising two layers and a hidden size of $64$.
The model was trained with a batch size of $32$ and a learning rate of $0.0001$ over the course of $200$ epochs.

\subsection{Results}

The experimental results are presented in Table~\ref{tab:main}. We find that: 
(1) Despite the high parameter efficiency of our LSTM model and its minimal training data requirements, its performance is comparable to that of other SOTA ORMs and PRMs.
(2) Our \method{} framework, which integrates LSTM with a PRM, consistently outperforms almost every other baseline model across all datasets and reasoning models.
(3) Our approach demonstrates significant improvements on two more challenging datasets, namely \texttt{MATH-500} and \texttt{CollegeMath}. We attribute the mediocre improvement on the \texttt{GSM8K} dataset to the low difficulty level of the dataset, which has resulted in data imbalance; a detailed discussion of this issue is provided in the following section.
\section{Discussion}

\subsection{Theorem Proving Strategy}
For each reasoning trajectory, we need to formalize and validate each step.
As a result, the quantity of theorems that require proving is quite significant.
As such, it is imperative to strike a balance between the success rate of theorem proving and the computational overhead involved.
Current endeavors in automated theorem proving typically address comprehensive mathematical problems, often employing tree search strategies and substantial search budgets.

Although employing a large search budget with thousands of searches can significantly increase the success rate of proving theorems for complete mathematical problems, the high computational cost of searching renders such approaches impractical for stepwise validation.
Therefore, we employed a computationally efficient setting by opting not to use the complex Monte Carlo tree search (MCTS) strategy employed by the DeepSeek-Prover-V1.5 \citep{xin2024deepseek15}.
We found that DeepSeek-Prover-V1.5 with a sample budget of 16 is sufficient to prove over 80\% of the statements in \bench described below.
Therefore, we adopted DeepSeek-Prover-V1.5 with a sample budget of 16 but without MCTS as the default theorem-proving strategy for other experiments. For additional evaluation on \bench, please refer to Appendix~\ref{sec:bench-eval}. 

\subsection{The \bench{} Dataset}

\begin{table}[t]
    \centering
\begin{scriptsize}
\begin{tabular}{*{7}{c}}
  \toprule
  & \textbf{Geo} & \textbf{Num} & \textbf{Alg} & \textbf{Comb} & \textbf{Oth} & \textbf{Total} \\
  \midrule
\textbf{Frequency} & 873 & 11515 & 5525 & 9414 & 3482 & 30809 \\
\textbf{S Length} & 147.5 & 79.5 & 107.8 & 125.0 & 112.8 & 104.2 \\
\textbf{P Length} & 51.1 & 36.4 & 39.8 & 49.8 & 25.5 & 41.0 \\
\textbf{Proof Rate} & 72.3 & 82.1 & 81.7 & 81.4 & 79.1 & 81.2 \\
  \bottomrule
\end{tabular}
\end{scriptsize}
    \caption{Statistical information of the \bench{} dataset, including category distributions, statement lengths, proof lengths and proof rates. The abbreviations Geo, Num, Alg, Comb, Oth refer to Geometry, Number Theory, Algebra, Combinatorics, Others, respectively.}
    \label{tab:dataset-stat}
\end{table}

\noindent\textbf{Construction of \bench}
We randomly sampled $500$ problems from the training part of \texttt{MATH} dataset and employed Llama3.1 as the reasoning model to generate reasoning trajectories, subsequently passing these trajectories to the auto-formalization pipeline.
The $30,809$ theorems generated from this auto-formalization process were designated as a benchmark for ``step correctness theorem proving'', denoted as \bench \footnote{Since these theorems were obtained through auto-formalization, they may not necessarily be provable.}.

Following common mathematical competition classification methods \citep{minif2f}, we categorized the Lean 4 statements into four types: geometry, number theory, algebra, and combinatorics.
We employed the LLM-as-a-judge approach using the GPT-4o-mini model to classify the FormalStep dataset.
The statistics in Table~\ref{tab:dataset-stat} reveal that while previous automated theorem proving works in Lean (e.g., miniF2F \citep{minif2f}, FIMO \citep{liu2023fimo}) primarily focused on Number Theory and Algebra problems, our dataset contains a substantial proportion of Combinatorics statements and a smaller but notable portion of Geometry problems.
Notably, Geometry and Combinatorics statements exhibit significantly longer statement lengths and proof lengths, highlighting the inherent challenges in these categories.

\subsection{Auto-Formalization of Single Steps}

Our experiment indicates that the statement generated by the auto-formalization process can encompass not only numerical computations and solving systems of equations --- tasks that can be easily tackled by general-purpose programming languages such as Python --- but also verify specific mathematical properties that are frequently overlooked by general-purpose programming languages.

These mathematical properties include whether a specific condition is sufficient or necessary, or properties like ``an integer is divisible by three if and only if the sum of its digits is divisible by three''.
This highlights the advantages of employing formal verification, as utilizing a domain-specific language like Lean facilitates a more efficient expression of mathematical concepts.
Additional examples of auto-formalization are provided in Appendix~\ref{sec:example}.
For additional quality evaluations regarding auto-formalization, please refer to Appendix~\ref{sec:quality-eval}.

\begin{figure*}[t]
    \centering
    \begin{subfigure}{0.24\textwidth}
        \centering
        \includegraphics[width=\linewidth]{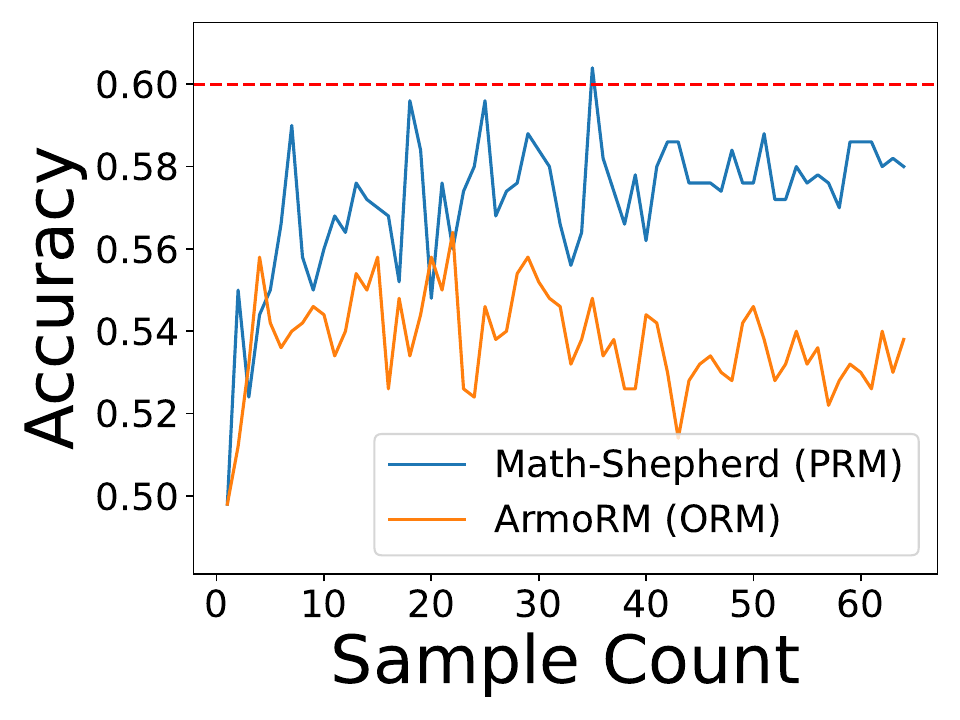}
        \caption{Llama 3.1}
    \end{subfigure}\hfill
    \begin{subfigure}{0.24\textwidth}
        \centering
        \includegraphics[width=\linewidth]{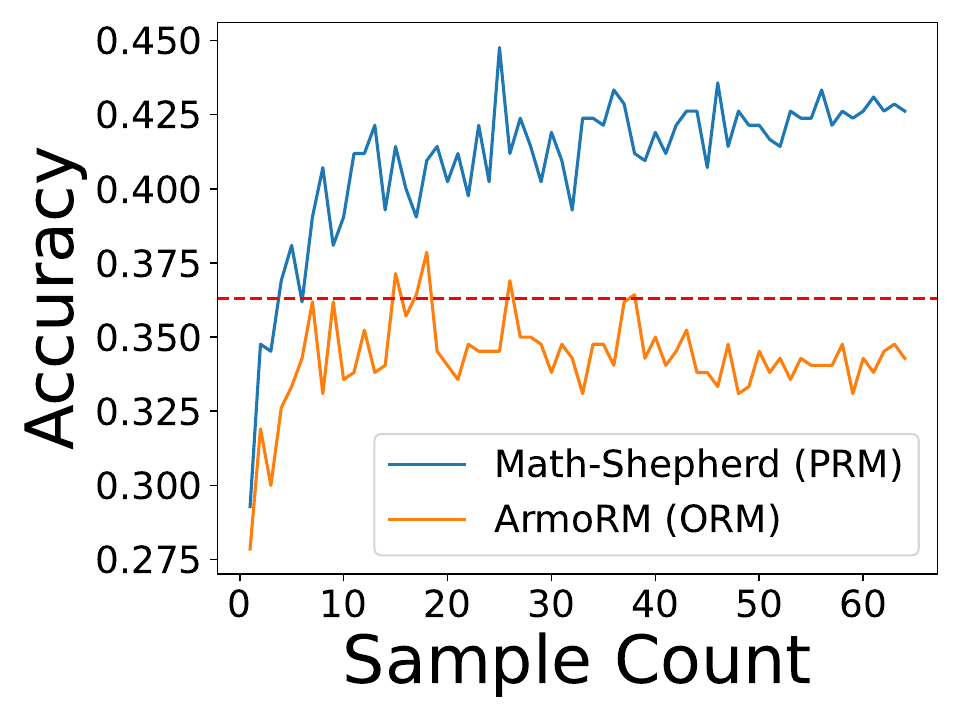}
        \caption{Llama 3.0}
    \end{subfigure}\hfill
    \begin{subfigure}{0.24\textwidth}
        \centering
        \includegraphics[width=\linewidth]{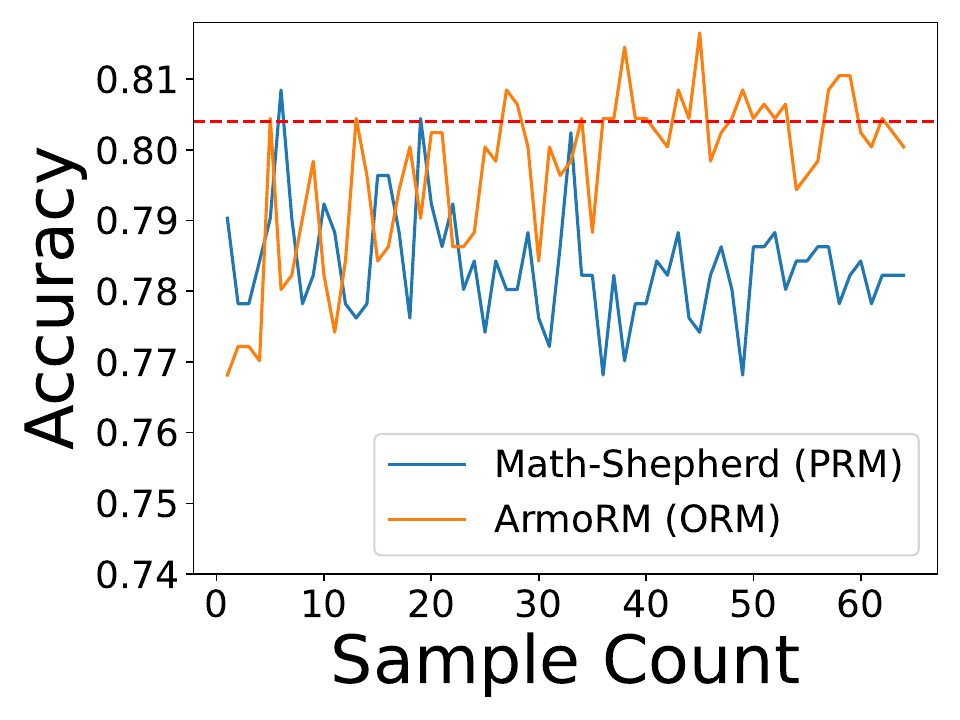}
        \caption{GPT-4o}
    \end{subfigure}\hfill
    \begin{subfigure}{0.24\textwidth}
        \centering
        \includegraphics[width=\linewidth]{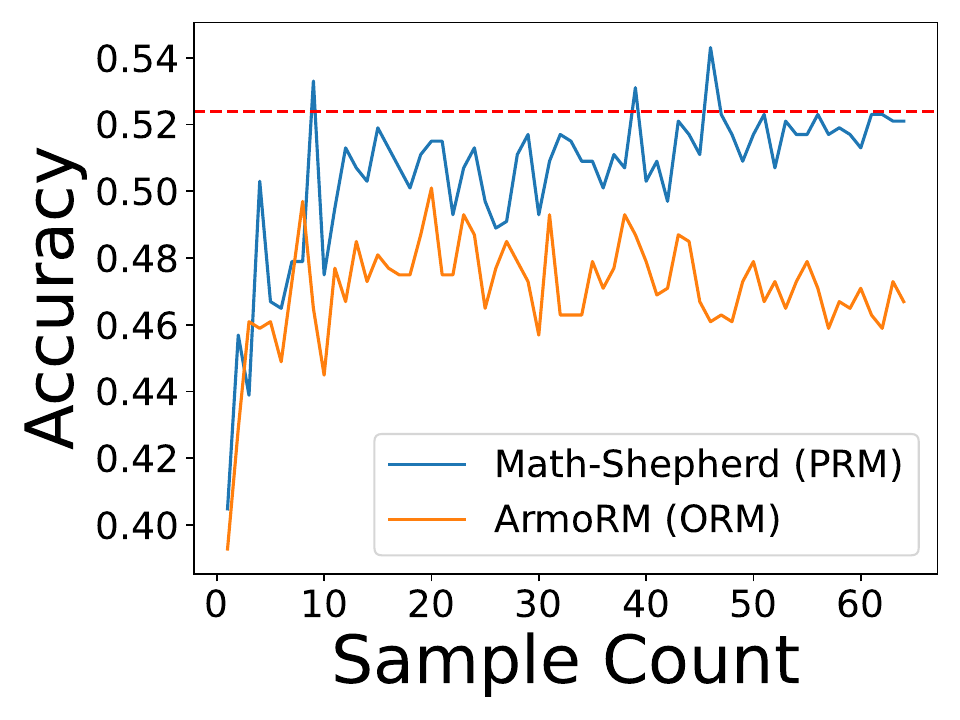}
        \caption{DeekSeek-Math}
    \end{subfigure}
    \caption{Scaling Best-of-N with Math-Shephed PRM and ArmoRM ORM. The red dashed line indicates the accuracy of \method{} using Best-of-N@5, while the plot demonstrates the variations in the accruacy of the selection of Best-of-N as N increases when utilizing the ORM or the PRM.}
    \label{fig:prm_scaling}
\end{figure*}

\subsection{Train \& Inference Cost}

In our analysis, we find that our method demonstrates high data efficiency. 
About $2,000$ reasoning trajectories sampled from $500$ questions, are generally sufficient to train an effective LSTM aggregator, which is significantly less than the \texttt{PRM800K} dataset, which contains approximately 800k step-level labels across 75k solutions \citep{lightman2024lets}, as well as the \texttt{Math-Shepherd} datasets, which comprise over 445k reasoning strategies \citep{wang2024mathshepherd}, which highlights the data efficiency of our proposed methodology.
We hypothesize that this discrepancy arises because the LSTM aggregator focuses on pattern recognition tasks centered on state sequences rather than directly processing natural language.
We also find that training an effective LSTM becomes challenging when there is a disproportionate ratio of correct to incorrect samples within the reasoning trajectories.
For instance, the four tested models are approaching saturation in the \texttt{GSM8K} dataset, resulting in a minimal presence of incorrect samples in the training set.
Consequently, the LSTM model struggled to learn the patterns of incorrect examples effectively, which may account for the relatively modest performance improvements observed with our approach on the \texttt{GSM8K} dataset.

Regarding inference costs, we recognize that our methodology requires more computational resources.
Specifically, each inference step entails a maximum of $3$ auto-formalization attempts and up to $16$ theorem proving attempts, with each attempt necessitating a query to an LLM.
During the curation of \bench, our analysis revealed that, on average, approximately $1.02$ automated formalization attempts and $6.97$ automated theorem proving attempts are conducted.
If we reduce the maximum number of ATP attempts from $16$ to $8$ --- while still achieving a proof success rate exceeding 80\% --- the average number of theorem proving attempts can be further decreased to $2.67$.
This suggests that the LLM queries of our single-step validation is approximately 4--8x greater than that of PRMs.

We increase the sample count for other RMs to compare two scaling strategies: increasing the sample count with a weaker RM or increasing computational resources to verify each sample.
The results presented in Figure~\ref{fig:prm_scaling} indicate that the scaling patterns during the testing phase vary among different LLMs.
For stronger models such as GPT-4o and Llama 3.1, it is advantageous to allocate additional computational resources during the verification process to achieve enhanced performance.
Conversely, for less powerful models like Llama 3.0, the quality of the sampled reasoning trajectories is subpar.
Therefore, increasing the sample count with a weaker yet more cost-effective RM may remain an effective strategy for these models.
\subsection{Synergistic Effect between PRM and Formal Step Verifier}

\begin{table}[t]
    \centering
\begin{footnotesize}
\begin{tabular}{*{5}{c}}
  \toprule
  \textbf{MATH-500 Acc} & \multicolumn{1}{c}{\textbf{L 3.1}} & \multicolumn{1}{c}{\textbf{L 3.0}} & \multicolumn{1}{c}{\textbf{GPT-4o}} & \multicolumn{1}{c}{\textbf{DSM}} \\
  \midrule
  Skywork ($\text{ORM}_1$) & 48.9 & 30.5 & 76.7 & 43.6 \\
  ArmoRM ($\text{ORM}_2$) & 55.1 & 32.3 & 79.3 & 48.6 \\
  Shepherd ($\text{PRM}_1$) & 58.1 & 34.7 & \underline{79.8} & 49.7 \\
  RLHFlow ($\text{PRM}_2$) & 51.7 & 29.6 & 78.7 & 44.4 \\
  \midrule
  $\text{ORM}_1 \oplus \text{PRM}_1$ & 58.6 & 35.2 & 79.6 & 49.7 \\
  $\text{ORM}_1 \oplus \text{PRM}_2$ & 55.9 & 34.1 & 79.3 & 47.6 \\
  $\text{ORM}_2 \oplus \text{PRM}_1$ & \underline{59.6} & \underline{35.6} & \underline{79.8} & \underline{50.1} \\
  $\text{ORM}_2 \oplus \text{PRM}_2$ & 57.1 & 33.8 & \underline{79.8} & 48.2 \\
  \midrule
  \textbf{LSTM (Ours)} & 55.1 & 33.0 & 78.9 & 48.2 \\
  \textbf{\method{} (Ours)} & \textbf{60.0} & \textbf{36.3} & \textbf{80.4} & \textbf{52.4} \\
  \bottomrule
\end{tabular}
\end{footnotesize}
    \caption{An ablation analysis that combines various ORMs and PRMs. The accuracy is assessed using the BoN@5 metric. The abbreviations L 3.1, L 3.0, GPT-4o, and DSM refer to Llama 3.1, Llama 3.0, GPT-4o, and Deepseek-Math model, respectively.}
    \label{tab:ablation}
\end{table}

Our main experiment indicates a synergistic effect between the PRM and our formal step verifier.
This observation prompts us to explore the potential existence of a similar synergistic effect between PRMs and ORMs.
We employed a similar strategy to integrate the PRM and the ORM.
The integration method employed aligns with our \method{} approach, wherein both models independently evaluate the same reasoning trajectory.
Following this evaluation, a coefficient is applied to combine the scores, resulting in a comprehensive assessment of the reasoning trajectory.
We employ the same metric, Best-of-N@5, to assess performance.

The results of the ablation experiments are presented in the Table~\ref{tab:ablation}.
The results indicate that PRMs and ORMs do benefit from model ensembling; however, the improvement is not as substantial as that achieved by \method.
Although the performance of our LSTM is comparable to that of an ORM, \method{} consistently outperforms ensemble models of all evaluated combinations, notably achieving a 2.3\% (50.1\% $\rightarrow$ 52.4\%) increase in the performance of DeepSeek-Math model.
We posit that the synergistic effect between the PRM and our formal step verifier is grounded in the complementarity of prospective and retrospective, as well as formal and informal, verification methods.

\section{Conclusion}

In this paper, we introduce a retrospective step-aware formal mathematical verification framework, termed \method{}, which utilizes auto-formalization and automated theorem proving to assign one of four distinct states to each step within the reasoning trajectory of LLMs when addressing mathematical problems.
To the best of our knowledge, this is the first approach to employ formal mathematical language Lean 4 for validating the correctness of LLM generated mathematical reasoning expressed in natural language.
The formal mathematical proofs offer interpretable evidence for the correctness of natural-language reasoning steps.
A benchmark consisting of $30,809$ formal statements, referred to as \bench, will be released to facilitate auto-formalization and automated theorem proving in low-compute environments.
Extensive experiments conducted across various LLMs and mathematical datasets illustrate the effectiveness of our methods and highlight the potential role of formal mathematical language in enhancing LLM reasoning.
\section*{Limitations}

The correlation between the output of the Lean 4 REPL and the final evaluation scores appears to be relatively indirect.
Furthermore, the scoring mechanism based on LSTM neural network parameters still lacks perfect interpretability.
While verification is conducted at a stepwise level, our method does not provide a precise reward score for each step due to the noise introduced by the current limitations of both auto-formalization and automated theorem proving.
To address this challenge, we propose identifying critical proof steps by examining the correlation between intermediate verification states and the final outcome through systematic analysis of the compact LSTM architecture's parameters and decision mechanisms.

Theoretically, our proposed \method{} is capable of fulfilling dual roles: it can function as a verifier during the testing phase and as a reward model during the reinforcement training process.
However, the increased computational overhead may present frictions for the direct application of the current method in reinforcement learning scenarios.
Moreover, as shown in Figure~\ref{fig:prm_scaling}, in some instances, when verification costs are similar, the performance gains may not be significant.
Therefore, we identify the following two points as areas for future work: to alleviate the computational resource demands of the existing method, and to integrate formal language verification into the reinforcement learning pipeline.
\section*{Acknowledgments}

This paper is partially supported by grants from the National Key Research and Development Program of China with Grant No. 2023YFC3341203 and the National Natural Science Foundation of China (NSFC Grant Number 62276002).

\bibliography{custom}

\appendix

\newpage

\section{Prompt for Auto-Formalization}
\label{sec:prompt}

The following prompt is utilized to instruct an LLM in the task of auto-formalization.
It consists of four components: a task description, detailed instructions, a comparison of key differences between Lean 3 and Lean 4, and a selection of manually curated few-shot examples. The steps included in the few-shot examples are derived from \texttt{PRM800K} dataset.
We observed that during auto-formalization, LLMs may conflate the syntactical elements of Lean 3 with those of Lean 4.
To address this issue, we revised the prompts by incorporating key syntactical distinctions between Lean 3 and Lean 4, thereby guiding the model to generate outputs that are consistent with Lean 4.

\definecolor{keywordcolor}{rgb}{0,0,0}   
\definecolor{commentcolor}{rgb}{0,0,0}   
\definecolor{symbolcolor}{rgb}{0,0,0}    
\definecolor{sortcolor}{rgb}{0,0,0}      
\definecolor{errorcolor}{rgb}{0,0,0}           
\definecolor{stringcolor}{rgb}{0,0,0}    
\begin{lstlisting}[language=lean,basicstyle=\tiny\ttfamily,breaklines=true,frame=single]
Given a question and the steps to answer it, you need to determine whether the final step of the answer may involve a hallucination that requires theorem proving in Lean 4.
* If the step is simple and intuitive, and you are confident that it does not need verification, please answer False.
* However, you need to verify ** ALL NUMERICAL ** operations, no matter how simple or intuitive they may seem.
* If the step has a certain leap that is not very intuitive and may involve a hallucination, please provide a Lean theorem that can verify the step.
* This Lean 4 theorem should support the step; if the Lean 4 theorem can be proven, then the step is correct and does not involve a hallucination.
* Ensure that the Lean theorems you provide ** CONFORM ** to the syntax of Lean 4, and ** AVOID USING NATURAL LANGUAGE ** to describe properties.
* Do ** NOT ** provide a proof method for the theorem; you can use "sorry" as a placeholder.
* Output the formalized theorem of the final step or False, and do ** NOT ** output any other content or predict next step.
* Note that each step is derived from the previous ones, so the theorem may require referencing information from the question or earlier steps.

Note that Lean 4 is not backward compatible with Lean 3.
* Type constants are now in UpperCamelCase, for example, `Nat` and `List`. Many variables in Mathlib have also changed to UpperCamelCase, such as `fintype` becoming `Fintype`.
* Lambda expressions now use `=>` as the separator. For example, `fun x => x` is the identity function, instead of `λ x, x`.

### Question:
Let \[f(x) = \left\{
\begin{array}{cl} ax+3, &\text{ if }x>2, \\
x-5 &\text{ if } -2 \le x \le 2, \\
2x-b &\text{ if } x <-2.
\end{array}
\right.\]
Find $a+b$ if the piecewise function is continuous (which means that its graph can be drawn without lifting your pencil from the paper).

### Step to be verified:
For the piecewise function to be continuous, the cases must "meet" at $2$ and $-2$.
### Lean:
False

### Step to be verified:
For example, $ax+3$ and $x-5$ must be equal when $x=2$.
This implies $a(2)+3=2-5$, which we solve to get $2a=-6 \Rightarrow a=-3$.
### Lean:
```lean
theorem test
  (a x: ℝ)
  (h₀: a * x + 3 = x - 5)
  (h₁: x = 3):
  (a = (-3)) := by sorry
```

... (More steps have been omitted.)

### Question
<question>

### Steps that do not require verification:
<answer>
### Step to be verified:
<step>
### Lean:
\end{lstlisting}
\definecolor{keywordcolor}{rgb}{0.7, 0.1, 0.1}   
\definecolor{commentcolor}{rgb}{0.4, 0.4, 0.4}   
\definecolor{symbolcolor}{rgb}{0.0, 0.1, 0.6}    
\definecolor{sortcolor}{rgb}{0.1, 0.5, 0.1}      
\definecolor{errorcolor}{rgb}{1, 0, 0}           
\definecolor{stringcolor}{rgb}{0.5, 0.3, 0.2}    

\section{Prompt for LLM-as-a-Judge Evaluation}
\label{sec:evaluation_prompt}

The following prompt is utilized to instruct an LLM to evaluate the semantic alignment between the reasoning step and the auto-formalized Lean 4 statement.

\definecolor{keywordcolor}{rgb}{0,0,0}   
\definecolor{commentcolor}{rgb}{0,0,0}   
\definecolor{symbolcolor}{rgb}{0,0,0}    
\definecolor{sortcolor}{rgb}{0,0,0}      
\definecolor{errorcolor}{rgb}{0,0,0}           
\definecolor{stringcolor}{rgb}{0,0,0}    
\begin{lstlisting}[language=lean,basicstyle=\tiny\ttfamily,breaklines=true,frame=single]
Given a mathematical problem, its step-by-step reasoning chain, and a Lean 4 statement, you need to verify whether the Lean 4 statement corresponds to the final reasoning step.
If they are properly aligned, then proving the Lean 4 statement mathematically validates the correctness of the final step.
If they don’t match, the Lean 4 statement is irrelevant to the final step’s correctness.
The three categories are ["good", "fair", "poor"].
Do ** NOT ** respond with any other characters.

### Problem:
<problem>

### Chain-of-thoughts:
<cot>

### Step to Verify:
<step>

### Lean 4 Statement:
<statement>

Which category does this statement fall into? Please respond with one of ["good", "fair", "poor"].
\end{lstlisting}
\definecolor{keywordcolor}{rgb}{0.7, 0.1, 0.1}   
\definecolor{commentcolor}{rgb}{0.4, 0.4, 0.4}   
\definecolor{symbolcolor}{rgb}{0.0, 0.1, 0.6}    
\definecolor{sortcolor}{rgb}{0.1, 0.5, 0.1}      
\definecolor{errorcolor}{rgb}{1, 0, 0}           
\definecolor{stringcolor}{rgb}{0.5, 0.3, 0.2}    

The following prompt is employed to instruct an LLM to classify Lean statements into one of four predefined categories.

\definecolor{keywordcolor}{rgb}{0,0,0}   
\definecolor{commentcolor}{rgb}{0,0,0}   
\definecolor{symbolcolor}{rgb}{0,0,0}    
\definecolor{sortcolor}{rgb}{0,0,0}      
\definecolor{errorcolor}{rgb}{0,0,0}           
\definecolor{stringcolor}{rgb}{0,0,0}    
\begin{lstlisting}[language=lean,basicstyle=\tiny\ttfamily,breaklines=true,frame=single]
Given a Lean 4 theorem, you need to identify its category.
The five categories are ["geometry", "number theory", "algebra", "combinatorics", "others"].
* Do ** NOT ** respond with any other characters.

Theorem:
<statement>

To which category does this theorem belong?
Your answer should be one of ["geometry", "number theory", "algebra", "combinatorics", "others"].
\end{lstlisting}
\definecolor{keywordcolor}{rgb}{0.7, 0.1, 0.1}   
\definecolor{commentcolor}{rgb}{0.4, 0.4, 0.4}   
\definecolor{symbolcolor}{rgb}{0.0, 0.1, 0.6}    
\definecolor{sortcolor}{rgb}{0.1, 0.5, 0.1}      
\definecolor{errorcolor}{rgb}{1, 0, 0}           
\definecolor{stringcolor}{rgb}{0.5, 0.3, 0.2}    

\section{Decomposition Specifications}
\label{sec:decomposition}

\noindent\textbf{Heuristic Rules} In this approach, we utilize periods or line breaks as delimiters to partition the reasoning process into independent steps.
However, we have observed that this simplistic rule may result in excessively fragmented steps and may also lead to erroneous decomposition.
For instance, a period could serve as a decimal point rather than a separator between sentences.

\noindent\textbf{LLM In-Context Learning} This approach involves utilizing an LLM to decompose the reasoning process into distinct, independent steps.
We randomly select accurate reasoning processes from the \texttt{PRM800K} training set, which consists of manually annotated stepwise data, and utilize the stepwise partitioning method derived from these responses as few-shot examples.
The model is directed to produce a JSON-formatted array with each element representing a string that corresponds to an independent step.
Below are the step decomposition few-shot prompts.

\begin{lstlisting}[basicstyle=\tiny\ttfamily,breaklines=true,frame=single]
Given a solution to a problem, you need to break down the solution into individual logical steps.
You have to answer the question in JSON format.
The answer must be an Array, where each element is a string representing a logically independent step.
You can only split the answer simply, and you cannot make any modifications to the answer, even if the answer is incorrect.
You must include the complete solution without any missing steps.
Here are some examples.

### Solution:
I want to find the smallest positive integer $X$ that satisfies the given conditions. I know that any multiple of 3 has the form $3k$ for some integer $k$, so $X$ has the form $3k + 2$. I also know that any multiple of 5 has the form $5n$ for some integer $n$, so the number with the same units digit as $X$ has the form $5n + 4$. Since the units digits of $X$ and $5n + 4$ are the same, I can equate them and get $3k + 2 \equiv 5n + 4 \pmod{10}$. This means that the difference between $3k + 2$ and $5n + 4$ is a multiple of 10. I can simplify this difference by subtracting 4 from both sides and get $3k - 2 \equiv 5n \pmod{10}$. Now I need to find the smallest values of $k$ and $n$ that make this equation true. I can try some values of $k$ and see what they imply for $n$. If $k = 0$, then $3k - 2 = -2$ and $5n = -2 + 10m$ for some integer $m$. This implies that $n = -0.4 + 2m$, which is not an integer. So $k = 0$ does not work. If $k = 1$, then $3k - 2 = 1$ and $5n = 1 + 10m$ for some integer $m$. This implies that $n = 0.2 + 2m$, which is also not an integer. So $k = 1$ does not work either. If $k = 2$, then $3k - 2 = 4$ and $5n = 4 + 10m$ for some integer $m$. This implies that $n = 0.8 + 2m$, which is again not an integer. So $k = 2$ does not work as well. If $k = 3$, then $3k - 2 = 7$ and $5n = 7 + 10m$ for some integer $m$. This implies that $n = 1.4 + 2m$, which is not an integer. So $k = 3$ does not work. If $k = 4$, then $3k - 2 = 10$ and $5n = 10 + 10m$ for some integer $m$. This implies that $n = 2 + 2m$, which is an integer. So $k = 4$ works. This means that the smallest value of $X$ is $3k + 2 = 3 \cdot 4 + 2 = 14$.
### Steps:
```json
[
    "I want to find the smallest positive integer $X$ that satisfies the given conditions.",
    "I know that any multiple of 3 has the form $3k$ for some integer $k$, so $X$ has the form $3k + 2$.",
    "I also know that any multiple of 5 has the form $5n$ for some integer $n$, so the number with the same units digit as $X$ has the form $5n + 4$.",
    "Since the units digits of $X$ and $5n + 4$ are the same, I can equate them and get $3k + 2 \\equiv 5n + 4 \\pmod{10}$.",
    "This means that the difference between $3k + 2$ and $5n + 4$ is a multiple of 10.",
    "I can simplify this difference by subtracting 4 from both sides and get $3k - 2 \\equiv 5n \\pmod{10}$.",
    "Now I need to find the smallest values of $k$ and $n$ that make this equation true.",
    "I can try some values of $k$ and see what they imply for $n$.",
    "If $k = 0$, then $3k - 2 = -2$ and $5n = -2 + 10m$ for some integer $m$.",
    "This implies that $n = -0.4 + 2m$, which is not an integer.",
    "So $k = 0$ does not work.",
    "If $k = 1$, then $3k - 2 = 1$ and $5n = 1 + 10m$ for some integer $m$.",
    "This implies that $n = 0.2 + 2m$, which is also not an integer.",
    "So $k = 1$ does not work either.",
    "If $k = 2$, then $3k - 2 = 4$ and $5n = 4 + 10m$ for some integer $m$.",
    "This implies that $n = 0.8 + 2m$, which is again not an integer.",
    "So $k = 2$ does not work as well.",
    "If $k = 3$, then $3k - 2 = 7$ and $5n = 7 + 10m$ for some integer $m$.",
    "This implies that $n = 1.4 + 2m$, which is not an integer.",
    "So $k = 3$ does not work.",
    "If $k = 4$, then $3k - 2 = 10$ and $5n = 10 + 10m$ for some integer $m$.",
    "This implies that $n = 2 + 2m$, which is an integer.",
    "So $k = 4$ works.",
    "This means that the smallest value of $X$ is $3k + 2 = 3 \\cdot 4 + 2 = 14$."
]
```

... (More examples have been omitted.)

### Solution:
<solution>
### Steps:
\end{lstlisting}

We prioritize the LLM in-context learning setting; if the JSON output is not parsable, we revert to the heuristic rules-based approach as a fallback.

\section{Score Ensembling Strategies}
\label{sec:ensembling}

The design of weighted multiplication was intentionally kept simple and intuitive while proving effective in practice.
The key motivation is that when either the PRM score or LSTM verification score approaches 0 (indicating likely errors), the combined score should similarly reflect low confidence.
During development, we empirically evaluated four ensemble variants:

\begin{itemize}
    \setlength{\itemsep}{0pt}
    \setlength{\parsep}{0pt}
    \setlength{\parskip}{0pt} 
    \item Weighted summation
    \item Weighted multiplication (our final choice)
    \item Max selection
    \item Min selection
\end{itemize}

\begin{table}[t]
    \centering
\begin{footnotesize}
\begin{tabular}{{l}{c}}
  \toprule
    & \textbf{MATH BoN@5} \\
  \midrule
    \textbf{LSTM only} & 55.1 \\
    \textbf{PRM only} & 55.2 \\
  \midrule
    \textbf{Weighted Sum} & 59.6 \\
    \textbf{Weighted Mul (Safe)} & 60.0 \\
    \textbf{Min} & 57.0 \\
    \textbf{Max} & 58.2 \\
  \bottomrule
\end{tabular}
\end{footnotesize}
    \caption{The ablation study results comparing score ensembling strategies on Llama3.1 (\texttt{MATH-500}).}
    \label{tab:ensembling}
\end{table}

The ablation results in Table~\ref{tab:ensembling} show that weighted multiplication achieved the best performance.
This aligns with our design principle that strong negative signals from either component should significantly impact the final score.

\section{Analysis of State Distributions}
\label{sec:distribution}

Each step has one of four potential states that indicate whether the particular step of the reasoning trajectory contain flaws.
Ideally, each step within a correct reasoning trajectory should correspond to either the state of ``Proof Successful'' or ``No Verification Required'', and a step classified under the state of ``Proof Failed'' signifies that this step has inherent flaws.
However, due to the noise introduced during the processes of auto-formalization and automated theorem proving, this is not true for all reasoning trajectories.

\begin{figure}[th]
  \centering
  \vspace{-0.8cm}
  \includegraphics[width=\columnwidth]{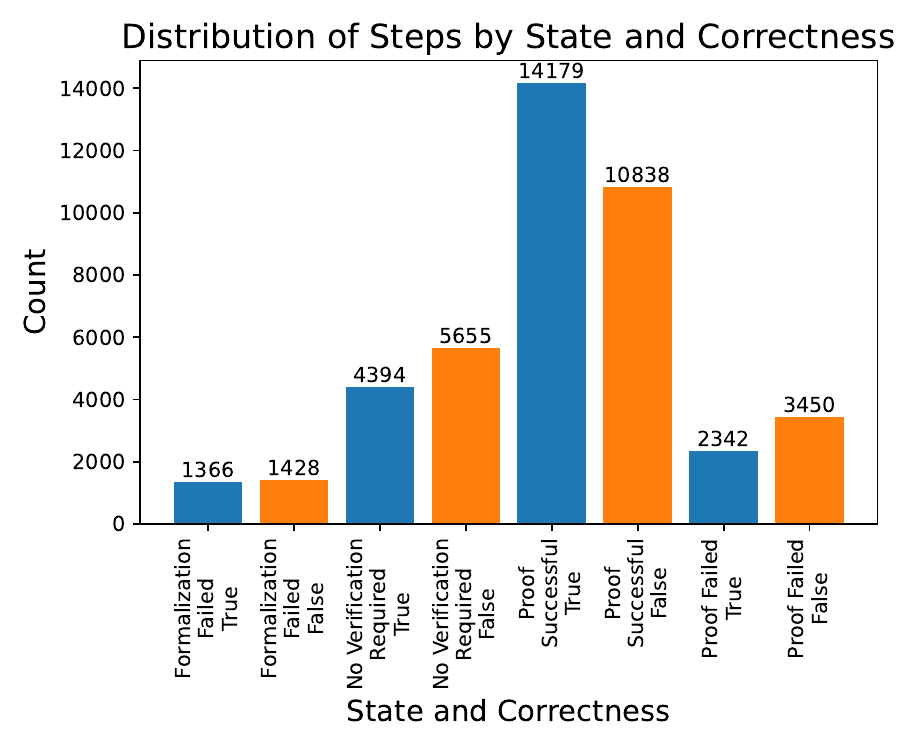}
  \caption{The distribution of steps by state and their correctness. Out of a total of $43,652$ steps, $30,809$ steps (72.2\%) were auto-formalized into valid Lean 4 statements. Among these, $25,017$ statements (81.2\%) were successfully proven by DeepSeek-Prover-V1.5.}
  \label{fig:ss_distribution}
  \vspace{-0.5cm}
\end{figure}

During the curation of \bench, we analyzed the distribution of states produced by our pipeline across these trajectories.
The statistical results are presented in Figure~\ref{fig:ss_distribution}.
Notably, the most common state identified is ``Proof Successful'', suggesting that the majority of the steps can be auto-formalized into valid Lean statements and be tackled by automated theorem provers.

We can observe that the likelihood of ultimately arriving at the correct answer for steps classified as ``Proof Successful'' is relatively high, at 56.7\%.
In contrast, the correctness for steps categorized under the ``Proof Failed'' state is significantly lower, at 40.4\%.
However, the noise introduced during the processes of auto-formalization and automated theorem proving complicates the direct assessment of the overall correctness of reasoning trajectories.

\section{Extra Examples of Auto-Formalization}
\label{sec:example}

The following example validates a theorem: an integer whose sum of digits is divisible by three can be inferred to be divisible by three itself.

\begin{lstlisting}[language=lean,basicstyle=\small\ttfamily,frame=single]
### Problem:
Let $N$ be the units digit of the number $21420N$. Which nonzero value of $N$ makes this number divisible by $6$?

### Reasoning Step by an LLM:
Divisibility by 3: A number is divisible by 3 if the sum of its digits is divisible by 3.

### Formal Statement & Proof:
theorem divisible_by_3
  (n N : ℕ)
  (h₀ : n = 2 + 1 + 4 + 2 + 0 + N)
  (h₁ : n % 3 = 0):
  (21420 + N) % 3 = 0 := by
  rw [h₀] at h₁
  simp_all [Nat.add_mod, Nat.mul_mod, Nat.mod_mod]
  <;> aesop
\end{lstlisting}

The following example is intended to demonstrate the sufficient and necessary conditions between equations.
This mathematical property may be challenging to articulate within the framework of general-purpose programming languages.
\newpage
\begin{lstlisting}[language=lean,basicstyle=\small\ttfamily,frame=single]
### Problem:
What real values of $x$ are not in the domain of

$f(x)=\frac{1}{|x^2+3x-4|+|x^2+9x+20|}$?

### Reasoning Step by an LLM:
This means that the denominator will be equal to zero if and only if $x+4=0$, regardless of the values of $x-1$ and $x+5$.

### Formal Statement & Proof:
theorem test
  (x: ℝ):
  (|x + 4| * |x - 1| + |x + 4| * |x + 5| = 0) ↔ (x + 4 = 0) := by
  constructor
  intro h
  cases' le_total 0 (x + 4) with h₀ h₀ <;>
    cases' le_total 0 (x - 1) with h₁ h₁ <;>
      cases' le_total 0 (x + 5) with h₂ h₂ <;>
        simp_all only [abs_of_nonneg, abs_of_nonpos, add_left_neg, add_right_neg,
          add_zero, zero_add, mul_zero, zero_mul, mul_one, mul_neg, neg_mul, neg_neg,
          mul_right_comm, mul_left_comm] <;>
        nlinarith
  intro h
  rw [h]
  simp
\end{lstlisting}

The following example illustrates that the Lean 4 language is capable of performing some tasks typically associated with general-purpose programming languages, such as numerical computation.

\begin{lstlisting}[language=lean,basicstyle=\small\ttfamily,frame=single]
### Problem:
A shirt's sale price is marked $\$14.40$, which is $60\%$ off the original price. How many dollars was the original price of the shirt?

### Reasoning Step by an LLM:
So, the original price of the shirt was $36.

### Formal Statement & Proof:
theorem test
  (sale_price original_price: ℝ)
  (h₀: sale_price = 14.40)
  (h₁: sale_price = 0.4 * original_price):
  (original_price = 36) := by
  rw [h₀] at h₁
  ring_nf at h₁
  linarith
\end{lstlisting}

\section{Additional Quality Evaluation Regarding Auto-formalization}
\label{sec:quality-eval}

Regarding accuracy (here defined as whether the auto-formalized statements comply with the Lean~4 syntax \citep{ying2024lean}), 72.2\% ($30,809$ out of $43,652$) of steps were successfully formalized into Lean 4-compliant statements when constructing the \bench.
Concerning consistency (here defined as semantic alignment between successfully formalized statements and original natural language statements  \citep{ying2024lean}), we employed GPT-4o-mini to evaluate our \bench{} using an LLM-as-a-Judge approach.
The complete prompt utilized in the LLM-as-a-Judge evalution process is available in Appendix~\ref{sec:evaluation_prompt}.

\begin{table}[h]
    \centering
\begin{footnotesize}
\begin{tabular}{*{4}{c}}
  \toprule
  \textbf{Good} & \textbf{Fair} & \textbf{Poor} & \textbf{Total} \\
  \midrule
    24,938 (80.9\%) & 138 (0.4\%) & 5733 (18.6\%) & 30,809 \\
  \bottomrule
\end{tabular}
\end{footnotesize}
    \caption{The evaluation results of semantic alignment on the \bench{} using the LLM-as-a-Judge method.}
    \label{tab:llm-as-a-judge}
\end{table}

The results in Table~\ref{tab:llm-as-a-judge} show 80.9\% of formalized theorems maintain good semantic alignment, which aligns with both recent work \footnote{Note that these recent studies report results for conventional auto-formalization tasks (i.e., translating problem statements), thus precluding direct comparison with our novel auto-formalization approach (i.e., translating individual solution steps).} (72\% for \texttt{FormL4} \citep{lu2024process} and 93.5\% for \texttt{Lean Workbook} \citep{ying2024lean}) and our manual observations during the design phase.
We use an LSTM aggregator rather than rejecting all unprovable steps outright, therefore our method should be robust to imperfect consistency.

\begin{figure}[t]
  \centering
  \includegraphics[width=\columnwidth]{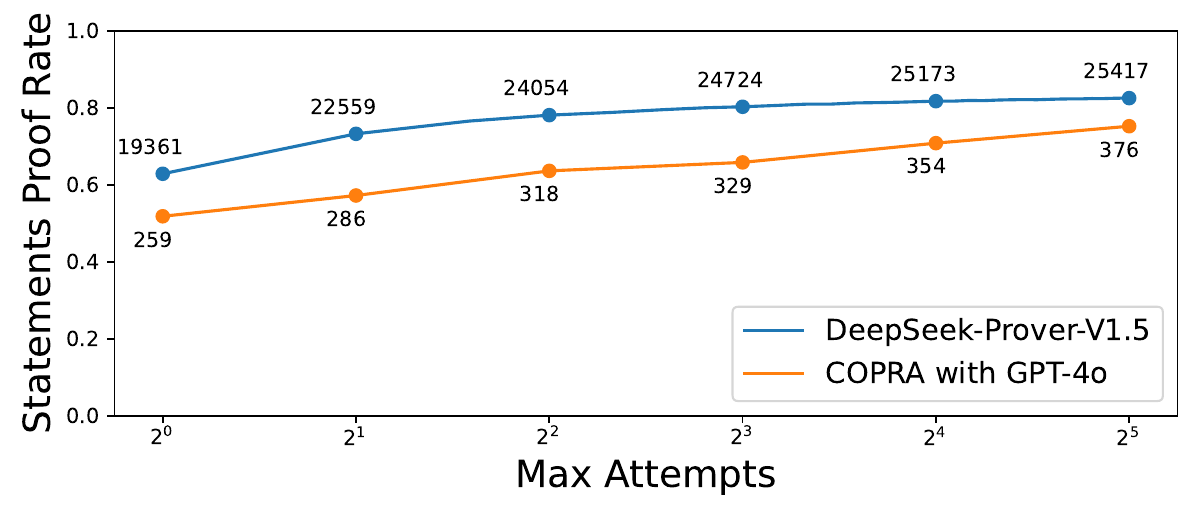}
  \caption{The proof rates of two SOTA theorem provers on \bench. The numbers indicate the total quantity of statements that have been successfully proved. Note that COPRA is significantly more resource-intensive due to its reliance on GPT-4o; therefore, it is tested on a randomly selected subset of $500$ statements, while DeepSeek-Prover-V1.5 is evaluated on full \bench, comprising a total of $30,809$ statements.}
  \label{fig:statement_proof_rate}
\end{figure}

\section{Additional Evaluation of \bench}
\label{sec:bench-eval}
We conducted experiments with DeepSeek-Prover-V1.5 and COPRA combined with GPT-4o under varying sample budgets, and the results are illustrated in Figure~\ref{fig:statement_proof_rate}.
Our results indicate that both provers are capable of effectively addressing the ATP task of proving single-step statements when provided with a substantial computational budget.
DeepSeek-Prover-V1.5 exhibits a significantly greater performance advantage over the agent-based COPRA in low-compute scenarios.

\end{document}